\documentclass[utf8]{frontiersSCNS} 
\usepackage{url,hyperref,lineno,microtype}
\usepackage[onehalfspacing]{setspace}
\usepackage{caption}
\usepackage[]{algorithm2e}

\def\firstAuthorLast{Senft {et~al.}} 
\def\Authors{Emmanuel Senft$^{1*}$, Michael Hagenow$^{2}$, Kevin Welsh$^{1}$, Robert Radwin$^{3}$, Michael Zinn$^{2}$, Michael Gleicher$^{1}$, and Bilge Mutlu $^{1}$}

\def\Address{
$^{1}$Department of Computer Sciences, University of Wisconsin--Madison, Madison, WI, USA\\
$^{2}$Department of Mechanical Engineering, University of Wisconsin--Madison, Madison, WI, USA\\
$^{3}$Department of Industrial System and Engineering, University of Wisconsin--Madison, Madison, WI, USA}

\def\corrAuthor{Emmanuel Senft}

\def\corrEmail{esenft@wisc.edu}

\AtBeginEnvironment{quote}{\itshape}

\newcommand{\method}{{\textit{Drawing Board}}}

\renewenvironment{quote}{%
   \list{}{%
    \leftmargin0.7cm   
    \rightmargin\leftmargin
   }
   \item\relax
}
{\endlist}

\graphicspath{{figs/}}

\begin{document}
\onecolumn
\firstpage{1}
\title{Task-Level Authoring for Remote Robot Teleoperation}

\author[\firstAuthorLast ]{\Authors} 
\address{} 
\correspondance{} 
\extraAuth{}

\maketitle

\begin{abstract}
Remote teleoperation of robots can broaden the reach of domain specialists across a wide range of industries such as home maintenance, health care, light manufacturing, and construction. However, current direct control methods are impractical, and existing tools for programming robot remotely have focused on users with significant robotic experience. Extending robot remote programming to end users, i.e., users who are experts in a domain but novices in robotics, requires tools that balance the rich features necessary for complex teleoperation tasks with ease of use. The primary challenge to usability is that novice users are unable to specify complete and robust task plans to allow a robot to perform duties autonomously, particularly in highly variable environments. Our solution is to allow operators to specify shorter sequences of high-level commands, which we call \textit{task-level authoring}, to create periods of variable robot autonomy. This approach allows inexperienced users to create robot behaviors in uncertain environments by interleaving exploration, specification of behaviors, and execution as separate steps. End users are able to break down the specification of tasks and adapt to the current needs of the interaction and environments, combining the reactivity of direct control to asynchronous operation. In this paper, we describe a prototype system contextualized in light manufacturing and its empirical validation in a user study where 18 participants with some programming experience were able to perform a variety of complex telemanipulation tasks with little training. Our results show that our approach allowed users to create flexible periods of autonomy and solve rich manipulation tasks. Furthermore, participants significantly preferred our system over comparative more direct interfaces, demonstrating the potential of our approach for enabling end users to effectively perform remote robot programming. 
\end{abstract}


\section{Introduction}
Effective \emph{teleoperation} of robots---broadly, a remote human controlling a robot at a distance \citep{niemeyer2016telerobotics}---is critical in scenarios where automation is impractical or undesirable. When a person operates a remote robot, they must acquire sufficient awareness of the robot's environment through sensors and displays, be able to make decisions about what the robot should do, provide directions (control) to the robot, and evaluate the outcomes of these operations. These challenges have been addressed with a wide range of interfaces that span a continuum of \emph{levels of autonomy} \citep{beer2014toward}, ranging from \textit{direct} control where the operator drives the moment-to-moment details of a robot's movements, to \textit{asynchronous} control, where operators send complex programs to the robot to execute autonomously, e.g., space exploration where robots receive programs for a day's worth of activities \citep{maxwell2005best}. The choices of level of control provide different trade-offs to address the goals of a specific scenario. In particular, longer-horizon control offers better robustness to communication issues and provides long periods of idle time for the operator while the robot is executing the commands. However, it also limits the opportunities for the human to react to unexpected situations during the program execution and requires significant huamn expertise to design robust behaviors and advanced sensing skills for the robot. On the other hand, more direct control allows operators to react quickly and easily to uncertainty, but demands constant attention from the operator, often relies on dedicated hardware, and requires a fast and stable connection to ensure that the tight real-time loop between the operator and the robot is maintained.

\begin{figure}[t]
  \centering
  \includegraphics[width=.6\linewidth]{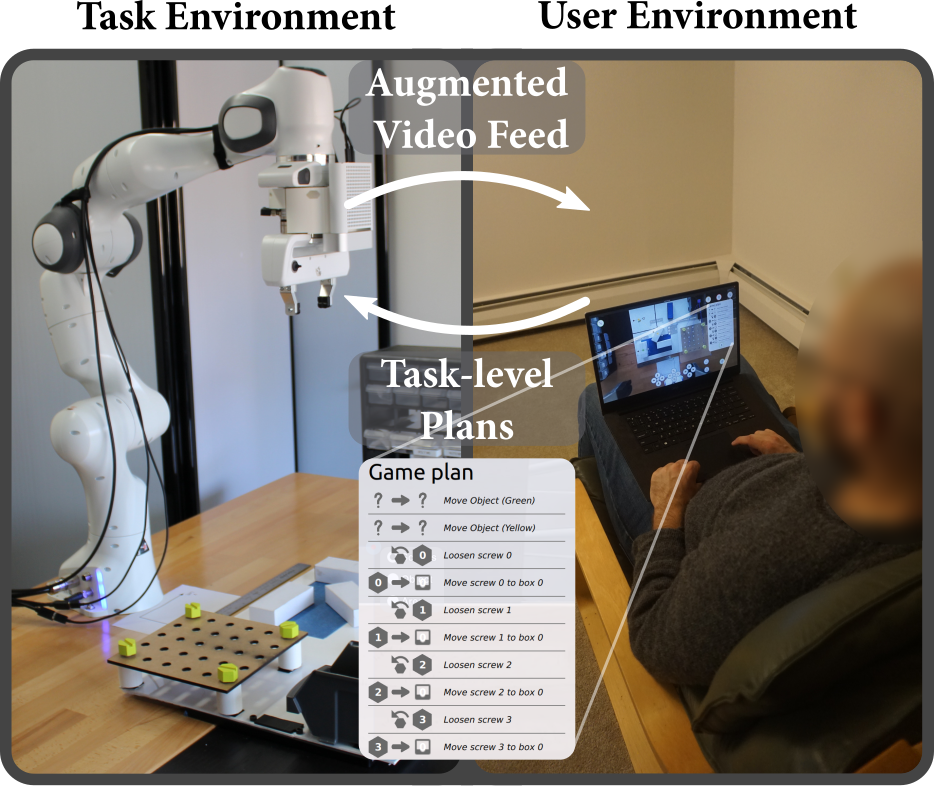}
  \caption{\method: task-level authoring for robot teleoperation. By watching a robot-centric augmented video feed and annotating it, novice users can acquire awareness about the robot's environment and send task-level plans to control a robot remotely.}
  \label{fig:teaser}
\end{figure}

Our goal is to provide effective telemanipulation for end-user applications, such as home care, light manufacturing, or construction. In such scenarios, high level robot autonomy of autonomy would be desirable, as this would reduce the operators' workload, however there remains situations where a fully autonomous behavior cannot be created. 
Users have domain knowledge, they can analyze the robot environment and determine appropriate actions for the robot, but they have no expertise in creating robot programs. Building a system supporting teleoperation for these novice users presents a number of challenges, the system needs to (1) be easy to use, (2) support active perception \citep{bajcsy1988active}, (3) support specification of robot behaviors adapted to the current state of the environment, and (4) allow for periods of autonomy.
As we will detail in Section \ref{sec:related}, current interfaces for teleoperation are often specifically tailored to highly trained operators or adopt a low level of autonomy. The former are not suited to novice users and the latter forces users to continuously provide inputs to the robot, reducing both the usability over extended periods of time and increasing the sensitivity to communication issues. %

Our key idea is to use \emph{task-level authoring} to enable the operator to control the robot by specifying semantically connected sequences of high-level (task-level) steps. This paradigm supports various lengths of program depending on available environment information, ranging from single actions to longer plans. For example, a robot might need to open a drawer with a specific label and empty it, however the robot does not have character recognition. The operator could use the robot to locate the appropriate drawer and then create a plan for the robot to open this specific drawer, remove all items in it, and then close it. Task-level authoring aims to offer more flexibility for the operator, allowing them both to specify long periods of autonomy when possible, but also have a more direct control when necessary to allow the operator to obtain the environmental awareness necessary to make longer plans.

We propose four principles to support effective telemanipulation by novices:
\begin{enumerate}
   \item \emph{Interleaving observation and planning:} the stepwise nature of manipulation tasks allows phases of observing the environment to gain awareness with phases of acting on that information. Execution occurs asynchronously, allowing it to be robust against communication problems and providing idle time to the user. Users can assess the state of the environment, devise a short plan for the robot, execute it, and the restart the process with the new state of the environment.
   \item \emph{Controlling the robot at the action level:} instead of controlling the robot motions, operators can select actions for the robot (e.g., pick-up, pull, or loosen). Such higher level of control allows participants to focus on the task that needs to be solved instead of the robot's kinemathics or workspace geometry.
   \item \emph{Providing a unified, augmented reality interface:} task specification can be accomplished from a viewpoint chosen by the user to be convenient, as part of their awareness gathering process. This process allows us to use a screen overlay-based augmented reality interface that aggregates all the required information for decision making on a single view also used to specify actions. This single view makes the programming easier for the users as all the important information are available in a single place. 
   \item \emph{Specifying actions graphically:} the augmented-reality interface allows for details of operations to be specified and verified graphically in context, simplifying the interface further. Additionally, such graphical specification allows easily to generalize a plan to a group of objects of the same type.
\end{enumerate}
We have prototyped these ideas in a system called \method{} after an artist's portable drawing board (Figure \ref{fig:teaser}) and evaluated it in a user study with 18 participants.
Our central contribution is to show that a task-level authoring approach can be applied to teleoperation to create a system that affords both ease-of-use and asynchronous operation. In our study, remote operators (students with limited programming knowledge) were able to perform complex tasks, gaining the benefits of asynchronous operation (robustness to delays and opportunities for longer periods of idle time) with the ease-of-use and reactivity of more direct interfaces. Participants---some literally on the other side of the world---were able to teloperate the robot with little training and preferred our system compared to interfaces not embodying our principles. These findings show how the core choice of task-level authoring is supported by specific interface and implementation designs, yielding a system that meets our goals, allowing end users to remotely create short period of autonomy for robots.

\section{Related work}
\label{sec:related}
Our work brings elements from the field of authoring and end-user programming to teleoperation.

\subsection{Teleoperation}
\label{sec:teleop}
Fundamentally teleoperation refers to human control of robot actions, typically done remotely (i.e., the human and the robot are not collocated and the human can only perceive the robot's environment through artificial sensors and displays) \citep{niemeyer2016telerobotics}. With teleoperation, the question of the appropriate level of autonomy is important, especially in the presence of delay and partial situational awareness \citep{niemeyer2016telerobotics,yanco2015analysis}. Levels of autonomy form a continuum between direct control and long-term programs:
\begin{enumerate}
    \item \emph{Direct control}: low-level control where the operator is manually controlling all actions of the robot in real-time (e.g., remote surgery \citep{marescaux2001transatlantic}, military \citep{yamauchi2004packbot}).
    \item \emph{Semi-autonomy}: the human operator intermittently controls robot actions where required and can parameterize higher-level actions that are executed autonomously by the robot (e.g., search-and-rescue, DARPA Robotics Challenge \citep{johnson2015team}).
    \item \emph{Teleprogramming}: operators create programs defining actions and reactions to changes in the environment for the robot to execute over longer period of time (e.g., Mars rovers programmed every day for a full day of autonomy \citep{norris2005science}).
\end{enumerate}

Direct control has seen widespread use in the aerospace, nuclear, military, and medical domains \citep{niemeyer2016telerobotics} as it allows operators to quickly react to new information. However, this type of teleoperation requires constant inputs from the operator and is highly sensitive to communications problems.  Researchers have explored various methods to address this communication challenge. One direction of research involves optimizing the communication channel itself to reduce delay and allow the operator to have quick feedback on their actions \citep{preusche2006robotics}. Another method, shared control, seeks to make the process more robust to human error through means such as virtual fixture methods, which support the operator in their direct manipulation task \citep{rosenberg1993virtual} or alternating phases of teleoperation and autonomous operation \citep{bohren2013pilot}. Finally, a third alternative uses a virtual model of the workspace to provide rapid feedback to the user from simulation while sending commands to the robot \citep{funda1992teleprogramming}. 

On the other end of the spectrum, traditional programming for autonomy and teleprogramming provides only limited feedback to the operator about the robot behavior. Operators need to have complete knowledge about the task including all required contingencies, to create dedicated programs for each task. These programs must be robust enough to run autonomously for hours without feedback. Furthermore, the robot needs to have the sensing capabilities to capture and analyze every relevant information in the environment. This highly autonomous control method is especially useful where there are large time delays between the robot and the operator which prevents the operator from intervening in real-time, such as when controlling a rover on Mars \citep{maxwell2005best}.

A semi-autonomous robot is a middle ground between these two extremes: it can execute short actions autonomously, but relies on the human operator to determine a plan of action and provide the correct parameters for these actions. The human (or team of humans) can use the robot to actively collect information about the environment, and provides near real-time inputs to the robot. The DARPA robotic challenge explores this space. In this case, the robot can run parameterized subroutines while multi-person teams of highly trained operators analyze data from the robot and control it at various abstraction-levels (from joint angle to locomotion goal), including situations with unstable communication channels \citep{johnson2015team}. These subroutines can be parameterized by selecting or moving virtual markers displaying the grasping pose \citep{kent2020leveraging}, robot joint position \citep{nakaoka2014development}, or using affordance templates \citep{hart2014affordance}. In a retrospective analysis, \cite{yanco2015analysis} highlight the training required for operating the robots during these trials, and reports that researchers should explore new interaction methods that could be used by first responders without extensive training.

One approach to simplify both awareness acquisition and control (two keys aspects in teleoperation) is to use monitor-based augmented reality---overlaying digital markers on views from the real world \citep{azuma1997survey}. For example, \cite{schmaus2019knowledge} present a system where an astronaut in the space station controlled a robot on earth using this technology. Their point-and-click interface presents the video feed from the head camera of a humanoid robot with outlines of the detected objects and menus around the video. When clicking on one of these objects, the system filters the actions that can be done on this object to propose only a small subset of possible actions to the operator. Similarly, \cite{chen2011view} propose a multi-touch interface when actions are assigned to gestures on a video feed displayed on a touch screen. This type of point-and-click or gesture interface allows the remote operator to gain awareness about the environment and simply select high-level actions for the robot to perform. Simulations can also be overlaid with markers that users can manipulate to specify the desired position of a robot or its end-effector \citep{hashimoto2011touchme,hart2014affordance,nakaoka2014development}. However, despite these advances, little work has been done to explore and evaluate interfaces that allow naive operators to actively acquire awareness about the environment and create longer plans consisting of multiple actions.

\smallskip

\subsection{Authoring}

In the context of robotics, the term \emph{authoring} refers to methods allowing end users to create defined robot behaviors \citep{datta2012robostudio,guerin2015framework,weintrop2018evaluating}. The general process starts with a design period where an initial behavior is created, then the robot can be deployed in the real world and its behavior tested and refined in additional programming steps if needed. When the desired requirements are achieved, the authoring process is finished and the robot is ready to be deployed to interact autonomously. Authoring differs from classic programming in its focus on end users with limited background in computer sciences and seeks to address questions of how can these users design, or \emph{author}, behaviors using modalities such as tangible interactions \citep{sefidgar2017situated,huang2017code3}, natural language \citep{walker2019neural}, augmented- or mixed-reality \citep{cao2019ghostar,peng2018roma,akan2011intuitive,gao2019pati}, visual programming environments \citep{glas2016human,paxton2017costar}, or a mixture of modalities \citep{huang2017code3,porfirio2019bodystorming}. \cite{steinmetz2018razer} describe task-level programming as parameterizing and sequencing predefined skills composed of primitives to solve a task at hand. Their approach combines this task-level programming and programming by demonstration \citep{billard2008survey} to create manipulation behaviors.

While promising, classic authoring methods suffer from two limitations when applied to remote robot control. First, the authoring process is often considered as a single design step creating a fully autonomous behavior \citep{cao2019v,perzylo2016intuitive}. This monolithic approach differs from teleoperation which assumes that human capabilities (sensory or cognitive) are available at runtime to help the robot successfully complete a task. Second, many authoring methods such as PATI \cite{gao2019pati} or COSTAR \cite{paxton2017costar} use modalities only available in situations where the human operator and the robot are collocated (e.g., kinesthetic teaching, tangible interfaces, or \textit{in-situ} mixed reality). For example, teach pendants---which are interfaces provided by manufacturers of industrial robots---are designed to be used next to the robot and often require the operator to manually move the robot. Consequently, while available to end users, such methods are not possible to use remotely. Our work is in the line of \cite{akan2011intuitive}, who used augmented reality to specify plans for a robotic arm. However, our premise is that to enable novice users to teleoperate robots, active perception (i.e., environment exploration) and behavior specification should be interleaved and coupled through a single simple interface, and that manipulation of graphic handles is a powerful way to specify parameters for actions.

\section{Design}
\label{sec:design}
\begin{figure}
 \centering
 \includegraphics[width=\textwidth]{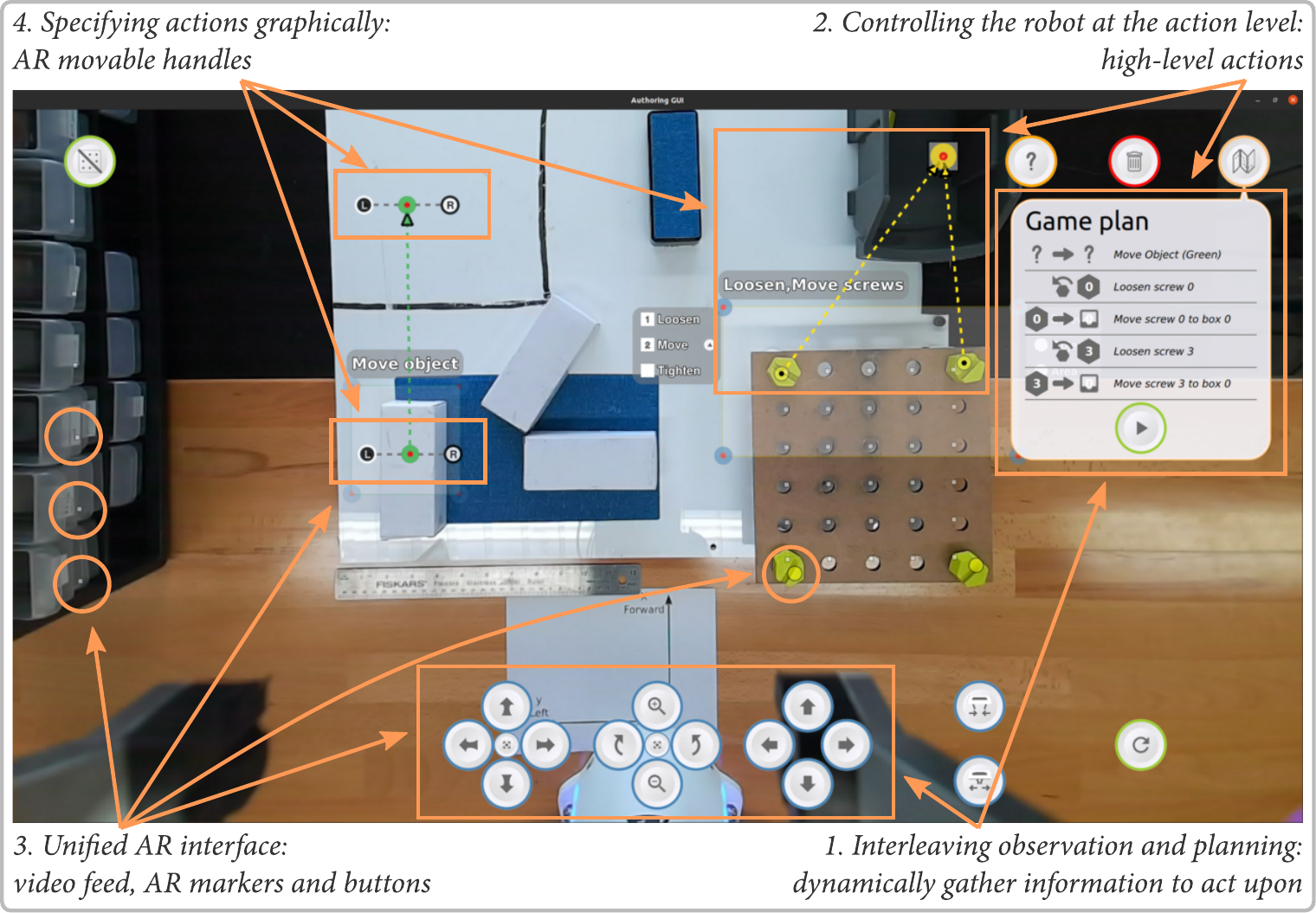}
 \caption{\method's interface demonstrating our design principles.}
 \label{fig:design}
\end{figure}

To allow non-expert users to control robots remotely, we propose a system rooted in task-level authoring which allows users to navigate the live environment and specify appropriate robot behaviors.
The following sections and Figure \ref{fig:design} detail the key concepts of the system topology.

\smallskip

\subsection{Interleaving observation and planning}

Specifying full execution plans for a robot would allow to reduce the operator workload during plan execution, but requires significant expertise in robotics and highly capable robots.
To allow end users to create adaptable periods of autonomy for the robot, we propose to use a task-level authoring approach. This approach simplifies the programming process by allowing the programmer to break tasks into sequences of high level actions based on what they observe at the moment. Users can chain together actions to create flexible periods of autonomy, adapted to their knowledge of the situation. For example, a set of actions may consist of grabbing a set of bolts in an area and moving them into a set grid pattern to fasten a structure. If the user is unsure what action is required next or if something unexpected occurs, the task-level authoring approach allows the user to explore the environment and create new programs based on the outcome of previous actions and new information.

Controlling robots at the task level creates a number of opportunities for end-user teleoperation; it allows the human to remain in the decision loop to provide necessary expertise, while maintaining an asynchronous workflow.
Such design allows end-users to alternate between observing the environment, specifying robot actions, and executing sequences of commands. Operators can specify short actions to explore the environment by moving the robot camera, acquiring awareness, and selecting an appropriate view point to author task plans. Then, once they have gathered enough information about the environment to know their next actions, they can schedule a longer plan consisting on multiple actions to solve the current part of the task. This process can be repeated as much as needed which allows for plans to be tailored to the current state of the environment. The asynchronous execution also provides robustness to communication instability. The inclusion of the operator in the control loop takes away the complexity of teleprogramming by having the human making complex perceptions and decisions. Thus, it keeps the benefits of direct control without the requirement of a tight and stable control loop and maintain the benefits of asynchronous control without requiring to create complex programs and plan ahead for unknown future.

\smallskip

\subsection{Controlling the robot at the action level}

As mentioned in Section \ref{sec:teleop}, teloperation levels of control covers a spectrum from direct control to teleprogramming. Direct control can afford ease of use when the user is provided with intuive input device \citep{rakita2018autonomous}, however it requires minute control from the operator and is very sensitive to delay. 

As shown in \cite{schmaus2019knowledge}, controlling a robot at the action level provides a number of advantages for teleoperation. First, as actions are executed using a local control loop, it allows to be robust to delays in communication. Second, it is intuitive for users, new operators can pick-up the system easily without requiring the user to possess any knowledge about robotics and control. 
Nevertheless, controlling solely at the action level suffers from some limitations. When using a single actions, even when users know what the robot should do over the next few actions, they have to specify an action, wait for it to be executed, specify the next action, and repeat, which can be suboptimal for the user. Additionally, similarly to any high-level control scheme, any action not in the robot vocabulary cannot be executed.

\smallskip

\subsection{Providing a unified, augmented reality interface}
\label{sec:awareness}
Similar to some other robot authoring interfaces \citep{schmaus2019knowledge,walker2019robot}, our approach uses augmented reality (AR) to simplify perception and action specification. The interface is composed of a unified monitor-based AR interface showing a live camera view of the robot's environment augmented with digital markers representing detected objects (see Figure \ref{fig:design}). The camera is mounted directly on the robot's end-effector for viewpoint flexibility and registration. The interface is overlaid with a canvas where the operator can design robot behaviors. This paradigm is consistent with research which shows that the most intuitive way to communicate information to an untrained operator is through vision \citep{yanco2004beyond}. More complex information such as the detected object pose and the environment point cloud are used to parameterize robot behavior behind the scenes, but hidden from the user's display. 

\smallskip

\subsection{Specifying actions graphically}
One challenge in designing an interface for novice end users is to simplify the specification of complex manipulations. In programming, classic ways to set parameters are through sliders and numbers. Numerical parameter-setting allows greater precision, but can be unintuitive for users. Instead, our interface design leverages graphical representations whenever possible and minimizes required user input.

Our interface uses visual and interactive representations, mapped onto the augmented video feed, that enable users to parameterize predefined actions by manipulating these graphical representations. For example, to move a known object to a known positions, the interface creates anchors that can be moved by the user. Then, the interface will display an arrow from the starting point in the video to the goal point, visually representing the action in context. The interface uses 2D affordances throughout, as this is consistent with the 2D representation in video. 6D locations are inferred from the 2D interface based on environment information. Additionally, through graphical localization, a series of actions on a specific object can be generalized to nearby objects of the same category.

Our interface design only exposes high-level actions to the user (e.g., move, tighten, pull). The local robot controller decomposes these high-level actions into series of lower-level actions and translates them into primitives to reach the desired robot behavior. The user only has to specify the minimum fields required to execute the task and graphical specification allows to specify multiple parameters at the same time and in an intuitive matter.
    
\section{Implementation}
\label{sec:implementation}
\subsection{System}
Following the considerations detailed in the previous section, we implemented \method, a prototype focused on enabling users with little programming experience to operate a robot remotely. The interface was designed to be served from a traditional laptop or desktop display and focuses on controlling a single robotic manipulator. Our implementation integrates a collaborative robot (Franka Emika Panda) outfitted with an ATI Axia80-M20 6-axis force torque sensor and a Microsoft Azure Kinect providing both a 2D image and 3D point cloud. The camera is placed at the end-effector to allow for the greatest flexibility in camera position. The components of the system communicate using ROS \citep{quigley2009ros} with logic nodes implemented in Python, the graphical user interface in QML, and the low-level control in C++.\footnote{Open-source code for our system implementation is available at \url{https://github.com/emmanuel-senft/authoring-ros/tree/study}.} 

For facilitating precise interaction with the environment, we implemented a hybrid controller to have more control over the forces applied by the robot when doing precise manipulation such as pulling a drawer. The hybrid control law follows an admittance architecture where interaction forces are measured from the force-torque sensor and resulting velocities are commanded in joint space via pseudo-inverse based inverse kinematics. 

We also leverage the Microsoft Azure Kinect depth sensor to observe the environment. Objects are first localized in the scene by feeding the color image to Detectron2 \citep{wu2019detectron2}, which provides a high fidelity binary pixel mask for each detected object. Once the object is localized, a GPU-accelerated Hough transform is used to register the known triangle mesh with each instance. This pipeline allows us to achieve 6D object pose estimation, which can then be used to provide the user with semantically correct actions as well as inform the robot motion plan. Our system also uses a number of predefined points of interest that represent the position of known static objects in the workspace. These known points are used as reference for the robot and to filter the position of objects detected by the live pose estimation pipeline.

\smallskip
\subsection{Workspace}

\begin{figure}[h]
  \centering
  \includegraphics[width=.7\linewidth]{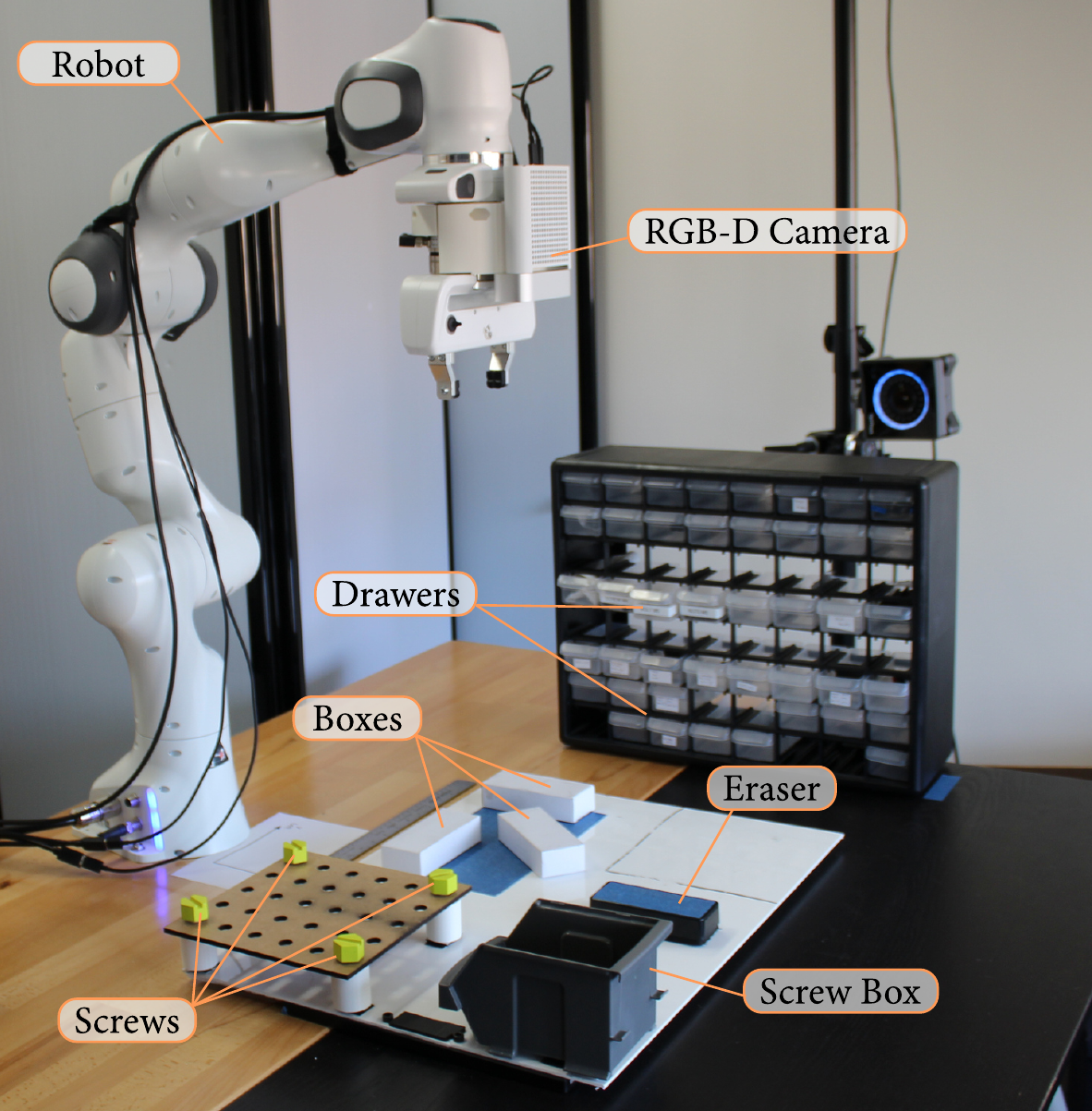}
  \caption{Workspace used for our implementation.}
  \label{fig:workspace}
\end{figure}

We applied our system to the workspace shown in Figure \ref{fig:workspace}. This workspace guided our implementation, but the system can be adapted to other tasks or interactive objects.
This workspace is composed of a number of drawers on the left of the robot with known positions. The middle of the workspace contains three white boxes above a blue area and a blue eraser, which are not detected by the robot object recognition system. The right part of the workspace contains a grid with holes and a screw box with known positions, and the grid can contain screws which are detected by the vision system.

The current prototype includes the following actions:
\begin{itemize}
    \item \emph{Pulling} and \emph{pushing} the drawers;
    \item \emph{Picking}, \emph{placing}, and \emph{moving} detected objects (e.g., screws) and undetected objects (e.g., boxes);
    \item \emph{Tightening} and \emph{loosening} the screws;
    \item \emph{Wiping} and area;
\end{itemize}
By having general actions such as \emph{pull} or \emph{move} our system can adapt easily to other objects or different locations and orientations for these objects.

\subsection{Interface}

The default interface layout shows the video feed augmented with markers showing the detected or known points of interest (see Figure \ref{fig:design}). The camera view is cropped to fill the full screen while showing clearly the robot's finger to allow users to know the gripper's status (open, closed, full).

\textbf{Direct Control:} At the bottom of the screen there are a number of buttons for direct control: 12 buttons allow the user to move the camera by a discrete increment in each of the 6 potential directions (5cm for the position buttons and $\pi/16$ radians for rotations), two buttons allow grasping and releasing, and a last button resets the robot to its homing position.

\textbf{Authoring:} To create task-level plans for the robots, users can annotate the augmented display to select actions applied to objects detected or parts of the environment. Users can click (or click-and-drag) on the screen to create selection areas to plan actions for the robot. Each selection area corresponds to one action or a set of actions on one type of object. Actions that can be parameterized (e.g., move actions) provide different types of handles that can be used to fully characterize the action.  Users can create multiple selection areas to schedule different types of actions, and the resulting plan is shown in the \emph{Game Plan} at the right of screen (see Figure \ref{fig:design}). Users can use this game plan to confirm that the interface interpreted the intentions correctly before sending the plan to the robot. During the execution, the user can monitor the robot progress in the task by watching the video feed and checking the progress in the plan. Video examples can be found at \url{https://osf.io/nd82j/}.

\begin{figure}[t]
  \centering
  \includegraphics[width=1\linewidth]{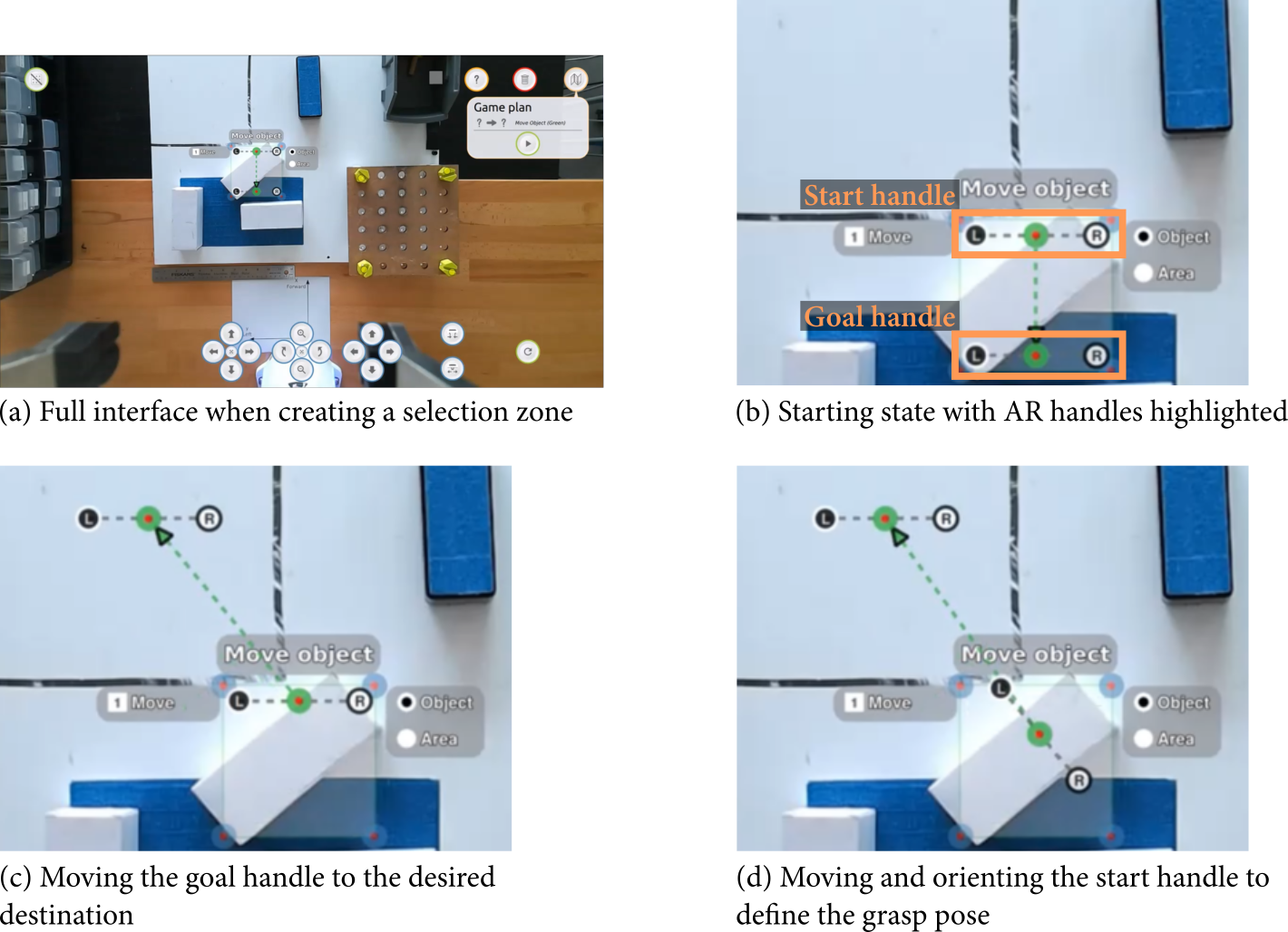}
  \caption{Example of parameterization of a moving unknown object action, with full interface for the initial state and zoom in.}
  \label{fig:walkthrough}
\end{figure}

\textbf{Interaction with undetected objects:} To pick and place an object not detected by the system, users can manipulate a start pose and a goal pose handles to specify the motion (see Figure \ref{fig:walkthrough}). These handles are composed of three connected points: the interaction point (grasping or releasing) as well as points representing the robot's fingers, and users can move the handle on the screen to change the interaction location, and rotate it to specify the end-effector orientation. This pixel value is then mapped into a 3D point in the camera frame using the Kinect's depth camera and converted in a point in space for robot. The orientation from the interface specifies the rotation on the vertical axis and consequently completely characterize a vertical tabletop grasp.

\textbf{Generalization to groups of objects:} When creating a selection area, the interface will select a default object to interact with based on the ones present in the area, but the type of object can be changed by clicking on radio buttons displaying the objects present in the area. Each object has a number of actions that can be executed on it (e.g., a screw can be tightened, loosened, or moved), and the user can select which action to apply and in what order by using numbered checkboxes. These actions will then be applied on each object of the selected type in the area (e.g., \textit{loosen} and \textit{move} all the screws in the area). 

\smallskip

\subsection{Backend}
The interface exposed the following high-level actions: move (both known and unknown objects), loosen, tighten, wipe, pull, and push. These high-level actions selected by users are grounded in the real world by finding the 6D pose of the interactions points in the user plan (either using the depth map from the Kinect or the location of points in a list of known objects). Each action is then hierarchically decomposed in a set of lower level actions (e.g., pick-up, view, place) and primitives (e.g., move to position, move to contact, grasps). For example, a \textit{move known object action} is decomposed first into a pick and a place actions, which are then decomposed into a multiple of primitives (move above grasping point, move to grasping point, grasp, move above grasping point, move above release point, move to release point, release, and finally move above releasing point). During execution, the robot will perform each of the parameterized primitive to complete the plan.

This method can be extended to new applications in three ways: (1) adding new objects to the image recognition and interface affordances, (2) by composing existing primitives to create new higher-level actions, and (3) if needed, by creating new primitives. The first two improvements could be made using graphical interface without having to code (e.g., using approaches similar to \cite{steinmetz2018razer}), however the third one would require actual code modification. This is similar to the current state-of-the-art cobots teach pendants: they expose a number of primitives that users can use to create behaviors, but any requirement not covered by the primitives (such as additional sensor-based interaction) would need code development to add the capability. Nevertheless, we could envisage a mixed system where creating new primitives and actions could be done locally, using learning from demonstration, and then exposed at a higher level to remote users using our interface.

\section{Evaluation}
We conducted an evaluation to assess the impact of the design principles of our system.
As mentioned in Section \ref{sec:design}, our task-level authoring system is based around four principles: (1) interleaving observation and planning, (2) controlling the robot at the action level, (3) providing a unified, augmented reality interface, (4) graphical specification of actions. For the sake of the evaluation, the \emph{unified AR interface} principle was not evaluated as too many different alternatives exist, however, we explored the three other axes. 
We conducted a 3$\times$1 within-participants study to explore three types of interfaces embodying or not our design principles: our task-level authoring interface (TLA), a point-and-click interface (PC) inspired from \cite{schmaus2019knowledge}, and finally a cartesian control interface (CC) as can be found traditionally on a cobot's teach pendant (e.g., PolyScope for the Universal Robots\footnote{\url{https://www.universal-robots.com}}) or recent work in teleoperation \citep{marturi2016towards}.

The CC condition does not use any of our design principles and serves as an alternative to kinesthetic teaching \citep{akgun2012trajectories} which cannot be applied due to the remote aspect and to direct control which would have required 6D input control on the user side. 
The PC condition only embodies the second design principle (\emph{controlling the robot at the action level}). It corresponds to a simpler version of our interface, where the robot has similar manipulation capabilities (e.g., pick-up objects, loosen or tighten screws, pull drawers) but where actions can only be specified one at a time and where parameters have to be set numerically (e.g., using use sliders to specify parameters such as angles). The last condition TLA is the interface described in Sections \ref{sec:design} and \ref{sec:implementation} and embodies all four of our principles. 

The evaluation took place over Zoom,\footnote{\url{https://zoom.us/}} a video conference platform, and we use the built-in remote screen control as a way to allow participants to control the robot from remote locations. We did not assess the latency inherent of such system, but estimated it around one second.

\subsection{Hypotheses}
Our evaluation uses the metrics $S$ for the task score, a performance measure; $A$ for robot autonomy, measured by both total and individual periods of autonomy; $U$ for usability, measured by the SUS scale \citep{brooke1996sus}; $P$ for user preference for the control method; and $W$ for workload, measured by the NASA Task-Load Index (NASA TLX) \citep{hart1988development}. Below, we describe our hypotheses and provide specific predictions for each measure. Subscripts denote study conditions (TLA, PC, and CC).

Our evaluation tested three hypotheses along the three evaluated design axes:
\begin{itemize}
    \item[$H_1$:] Task score, autonomy, usability, and user preference will be higher, and workload will be lower with high-level control (PC, TLA) than low-level control (CC). 
    \item[-] \emph{Prediction $P_{1a}$}: $S_{PC} > S_{CC}$, $A_{PC} > A_{CC}$, $U_{PC} > U_{CC}$, $P_{PC} > P_{CC}$, and  $W_{PC} < W_{CC}$.
    \item[-] \emph{Prediction $P_{1b}$}: $S_{TLA} > S_{CC}$,  $A_{TLA} > A_{CC}$, $U_{TLA} > U_{CC}$, $P_{TLA} > P_{CC}$, and  $W_{TLA} < W_{CC}$. 
    \item[$H_2$:] Autonomy and user preference will be higher, and the workload will be lower when users are able to interleave observation and planning (TLA) than when they are not able to (PC).
    \item[-] \emph{Prediction $P_{2a}$}: $A_{TLA} > A_{PC}$.
    \item[-] \emph{Prediction $P_{2b}$}: $W_{TLA} < W_{PC}$.
    \item[-] \emph{Prediction $P_{2c}$}: $P_{TLA} > P_{PC}$.
    \item[$H_3$:] Task score, usability, and user preference will be higher when users are able to parameterize actions graphically (TLA) than when they are not able to (PC).
    \item[-] \emph{Prediction $P_{3a}$}: $S_{TLA} > S_{PC}$.
    \item[-] \emph{Prediction $P_{3b}$}: $U_{TLA} > U_{PC}$.
    \item[-] \emph{Prediction $P_{3c}$}: $P_{TLA} > P_{PC}$.
\end{itemize}

$H1$ is based on the expectation that high-level action specification present in TLA and PC method automates away a large number of low-level actions that the user must specify in CC, which will save the user time and reduce the number of operations they must perform, thus their workload. $H_2$ is grounded in the expectation that the task planning offered by our system will be used by participants to create longer periods of autonomy, which should reduce the workload, and make the participants prefer the method. Finally, $H_3$ supposes that the graphical specification of actions will allow participants to specify action quicker (increasing their performance in the task), more easily (increasing the usability), and that participants will prefer this modality. 

\begin{figure}[!t]
  \centering
  \includegraphics[width=1\linewidth]{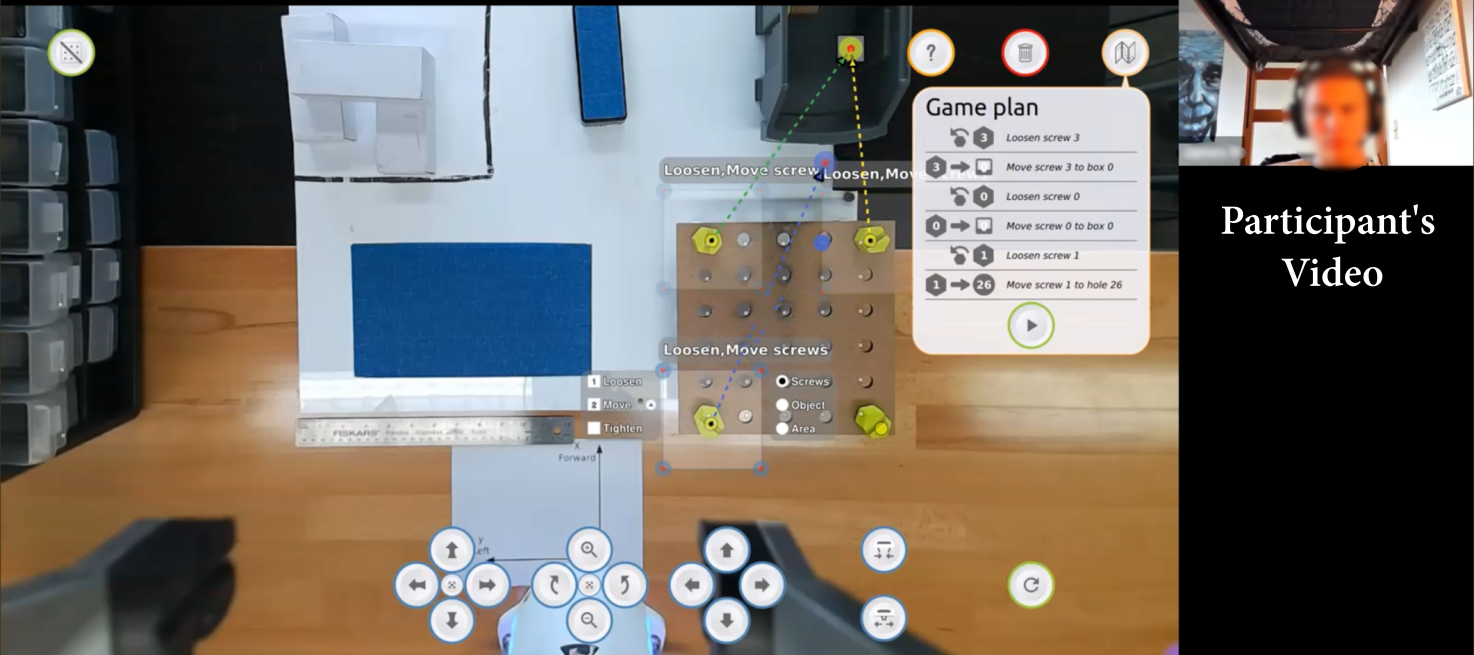}
  \caption{Example of a participant using \method{} to control the robot from his dorm bed.}
  \label{fig:participant}
\end{figure}

\smallskip
\subsection{Method}
\subsubsection{Participants}

 \begin{figure}[]
    \centering
    \includegraphics[width=\textwidth]{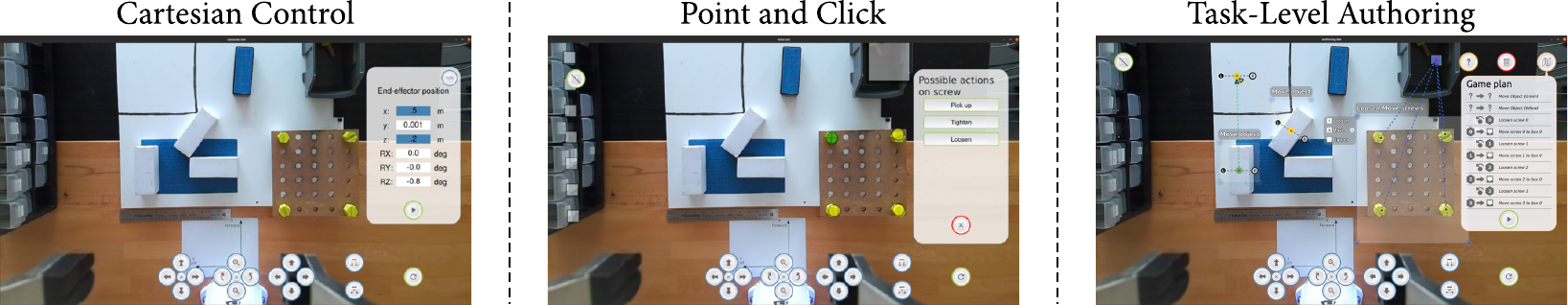}
    \caption{Interfaces used in the study. Left shows the Cartesian Control interface: the user specifies numerically the end-effector position. Middle shows the Point-and-Click interface: the user selects direct actions on objects. Right shows the Task-Level Authoring interface allowing users to remotely create task plans for the robot.}
    \label{fig:interfaces}
\end{figure}

We recruited 18 students enrolled in the Mechanical Engineering and Industrial and Systems Engineering departments at the university (3F/15M, age:  $M=19.6$, $SD=1.54$). We selected our participants from this population as they represent people with some exposure but little expertise in robotics (familiarity with robots $M=2.9$, $SD=1.2$ on a five-point scale---none, a little, some, moderate, a lot---and familiarity with programming $M=3$, $SD=0.6$). The procedure was approved by the university's Institutional Review Board and participants were compensated at the rate of \$15/hour. The study was designed to last 80 minutes and included around 45 minutes of robot operation. Since it was completed remotely, participants stayed in their daily environment as shown in Figure \ref{fig:participant} top right, where a participant controlled the robot from his dorm bed.

\smallskip
\subsubsection{Conditions}
In all conditions, the layout of the interface was the same. It showed the camera feed, overlaid with arrows for direct control, buttons to grasp and release and the reset button. The difference was in the type of command sent to the robot as well as the modalities provided to the user.
This study compared three conditions:

 \begin{enumerate}
     \item[CC] \textit{Cartesian control}: the user uses six text boxes showing the current Cartesian position of the end-effector ($x$, $y$, $z$, $rx$, $ry$, $rz$) (see Figure \ref{fig:interfaces}-left). These text boxes can be edited with the desired command and sent to the robot either after modifying a single dimension or multiple ones. 
     \item[PC] \textit{Point-and-Click}: the user is shown objects known by the system as markers overlayed on the video feed in an AR fashion (see Figure \ref{fig:interfaces}-center). The user can click on these markers or other parts of the view and is shown the different actions available on this object. Right clicking on an action allows to specify parameters, left clicking has the robot directly execute the action. 
     \item[TLA] \textit{Task-Level Authoring}: interface presented in Sections \ref{sec:design} and \ref{sec:implementation}, the user can annotate the video image to create actions associated to objects in the selection area and create task plans (see Figure \ref{fig:interfaces}-right).
 \end{enumerate}

To illustrate the difference between these conditions, we consider a move action on an unknown object (i.e., an object the operator can see, but the robot does not identify). With the CC condition, participants had to enter the 6D pose of a grasping point. Often, this process would be iterative, the operation would first have the robot approach the object by specifying a higher point, then correct the position and angle, then move to the grasping point. Then the operator had to press the grasp button, move to a dropping point (by specifying the 6D pose or using the camera control buttons), then press release. With the PC conditions, participants could click on the grasping location on the screen, parameterize the action with a grasp angle, execute the pick-up action, reset the robot, click on the destination location on screen and select the place action. With TLA, participant could click on the screen to create a section area, keep the \textit{move unknown object} action (the default one if no identified object was in the selection area), move the start and goal handles (as shown in Figure \ref{fig:walkthrough}), and finally press execute.

\smallskip

\subsubsection{Tasks}
As shown in Figure \ref{fig:interfaces}, the workspace has a number of drawers on the left, three white boxes on the bottom, an eraser at the top and four screws on the right. 

Participants had to complete a training task followed by four additional tasks:
\begin{enumerate}
    \item[]Task 0: \textbf{Training} move the angled white box to the top left area. 
    \item[]Task 1: \textbf{Pick-and-place} move the additional two boxes (at different orientations) to the top left area.
    \item[]Task 2: \textbf{Repeated actions} loosen the four screws from the grid and move them to the top-right gray box.
    \item[]Task 3: \textbf{Exploration} locate a specific drawer on the left, pull it, inspect its content and push it back.
    \item[]Task 4: \textbf{Continuous action} wipe the blue area with the eraser.
\end{enumerate}

These tasks were selected to represent different types of actions that a remote operator may need to complete. The first three pick-and-place actions free the area that need to be wiped, demonstrating workspace manipulation actions. The loosening and moving of the four screws represent repeated actions. The drawer inspection task combines two awareness acquisition actions: locating the drawer and inspecting its content, as well as two workspace manipulation actions: opening and closing the drawer. Finally, the wiping action represents a continuous action over an area, similar to cleaning a table or sanding a piece.

Of note, the three pick-and-place actions (task 0 and task 1) requires the human to specify manually the grasping and placing point as the robot does not detect the boxes by itself. And the exploration (task 3) requires the operator to gather information outside of the default field of view by moving the camera on the robot, read the labels on the drawers, locate the relevant drawer, open the drawer, look into the drawer, count the numbers of items, and finally close the drawer. To be able to complete this task autonomously, a robot would need to have optical character recognition capabilities and be able to detect and count any type of object present in the the drawers. Furthermore, as shown in Algorithm 1, if an operator wanted to design in a single step a program solving this task, the resulting program would require logic functions such as loops, conditional on sensors, functions within conditions, and loop breaking conditions. All these functionalities could be supported by more complex visual programming languages (such as Blockly\footnote{\url{https://developers.google.com/blockly}}) which requires more knowledge in programming. Such a program would also require more capabilities for the robot, more complex representation of the world (e.g., having a list of all drawers with positions to read the label from, and positions to inspect the content), and more complex programming languages. Instead, using a human-in-the-loop approach (through direct control or task-level authoring) allows to achieve the same outcome, but with much simpler robot capabilities and interfaces.

\begin{algorithm}[H]
 \For{Each drawers, d}{
    goToReadingPosition(d)\;
    l = readLabel(d)\;
    \eIf{l ==``Bolt M5''}
    {
        pull(d)\;
        goToInspectPosition(d)\;
        countObjects(d,``Bolt M5'')\;
        push(d)\;
        break\;
    }
    {continue\;}
    
 }
 \caption{Example of algorithm to solve the exploration task autonomously}
\end{algorithm}

\subsubsection{Procedure}
Participants joined a zoom call from their home or other daily environment. The study started with informed consent and a demographic questionnaire. Then participants were asked to watch a video introducing the robot and the workspace, followed by a second video introducing the tasks participants would need to complete\footnote{All the videos are available at \url{https://osf.io/nd82j/}.}. For each condition, participants first watched a two-minute video presenting the main modalities of the interface and demonstrating how to make a pick-and-place action. Then, participants had 15 minutes to complete as many of the tasks as possible. During the training, they could ask any questions to the experimenter, however in the four later tasks the experimenter was only able to answer the most simple questions (e.g., ``the screws are tightened down, right?'' but not ``how to move this object?''). 

The interaction with the robot stopped when participants reached 15 minutes or when they completed all the tasks. After this interaction, participants filled out NASA Task-Load Index (NASA TLX) \citep{hart1988development} and System Usability Scale (SUS) \citep{brooke1996sus} questionnaires before moving on to the next condition. The order of the conditions was counter-balanced and the study concluded with a semi-structured interview and a debriefing where participants could ask questions to the experimenter. Despite our best efforts, some participants created actions that could trigger the robot's safety locks (often due to excessive force being applied). In such situations, the timer was paused, the robot was restarted and the participant continued from where they stopped.

\smallskip
\subsubsection{Measurements}
We collected four types of quantitative data from the study. Task score, measured by how many tasks were fully or partially completed by the participants in the 15 minutes allocated per condition (with one point for the training and per task). Workload (i.e. how demanding it was to use the interface), measured by the NASA TLX. Usability (i.e. how intuitive the interface was), measured by SUS. Periods of autonomy, measured from period where the robot was moving continuously for more than 10 seconds (to be as inclusive as possible while not counting short periods that could barely be considered autonomous).
We measured the periods of autonomy with a mixture of logs from the system and video coding of the interaction recordings using the Elan software \citep{elan}.

In addition to the quantitative metrics, we collected qualitative impressions through the semi-structured interviews where we asked questions to the participants about their different experiences with the methods and which one they preferred.
    
\section{Results}
\begin{figure*}
    \centering
    \includegraphics[width=\textwidth]{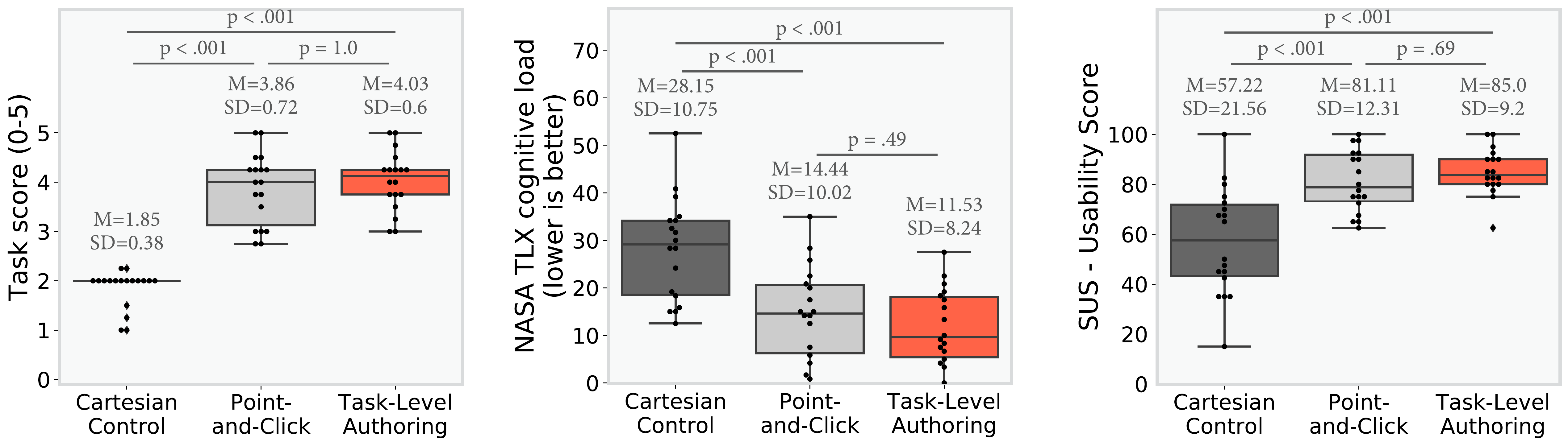}
    \caption{Study results: p-values are computed using post-hoc paired t-test adjusted with Bonferroni correction (n=18). Results show that both TLA and PC achieved a higher performance than CC (as shown by the task score), CC had a higher workload than both PC and TLA, and that both PC and TLA had a higher usability than CC. On the performance, workload, and usability no significant difference was observed between TLA and PC.}
    \label{fig:results}
\end{figure*}

Figure \ref{fig:results} and Figure \ref{fig:autonomy} present the quantitative results from the study. Results are first analyzed with a repeated measure ANOVA (corrected as needed using Greenhouse-Geisser), and then with post-hoc paired t-tests. A Bonferroni correction was directly applied to the p-values to protect against Type I error. For the periods of autonomy, as there was an unbalanced number of samples, we used ANOVA and Games-Howell post-hoc test.

\smallskip
\subsection{Score}

We observe significant impact of the condition on the score (sphericity was violated, Greenhouse-Geisser correction was used, $F(2,34)=115.53$, $p < .001$). Both the PC and the TLA interface achieved a score significantly higher than the CC (PC-CC: $t(17.0)=13.0$, $p < .001$, TLA-CC: $t(17.0)=-23.0$, $p < .001$). However, we do not observe a significant difference of score between the PC and the TLA interfaces ($t(17.0)=-1.0$, $p = 1.0$).

Additionally, we did not observe an impact of the order, indicating that there no significant learning effect ($F(2,34)=0.697671$, $p=.50$).

\smallskip
\subsection{Periods of autonomy}
\begin{figure*}
    \centering
    \includegraphics[width=.6\textwidth]{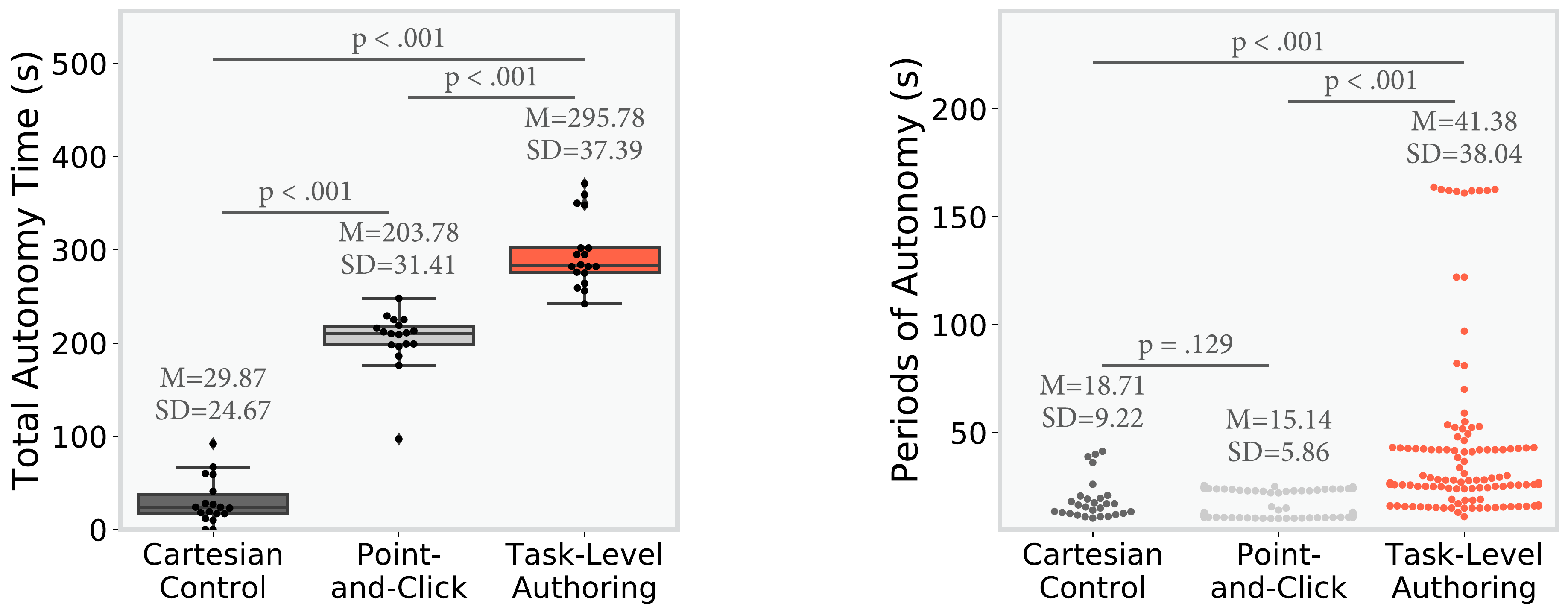}
    \caption{Autonomy results. Left shows the total autonomy time for each conditions (only periods of autonomy of more than 10 seconds are counted), p-values are computed using post-hoc paired t-test adjusted with Bonferroni correction (n=18). Right shows every single period of autonomy, p-values are computed using Games-Howell post-hoc to adjust for unequal sample size. }
    \label{fig:autonomy}
\end{figure*}

As shown in Figure \ref{fig:autonomy}, we observe a significant effect of the condition on the total autonomy time, $F(2,34)=297.45$, $p < .001$, and each condition was significantly different from the others, PC-CC: $t(17.0)=18.0$, $p < .001$, TLA-CC: $t(17.0)=-25.0$, $p < .001$, TLA-PC: $t(17.0)=-7.0$, $p < .001$). With the CC offering the least amount of autonomy time, PC in the middle, and TLA offering the most of autonomy time. It can be observed that in addition to have a higher total autonomy time, the TLA condition also led to longer individual periods of autonomy than the other conditions (one way ANOVA: $F(2,402)=60.87$, $p < .001$, Games-Howell post-hoc test PC-CC: Mean Difference$ = -3.57$ , $p = .129$, TLA-CC: Mean Difference $= -22.66$ , $p < .001$, TLA-PC: Mean Difference $= -26.24$, $p < .001$).

\smallskip
\subsection{Workload}
We observe significant effect of the condition on workload as measured by the NASA TLX, $F(2,34)=46.29$, $p < .001$. Both the PC and the TLA interface imposed a workload significantly lower than the CC (PC-CC: $t(17.0)=-9.0$, $p < .001$, TLA-CC: $t(17.0)=8.0$, $p < .001$). However, we do not observe a significant difference of workload between the PC and the TLA interfaces ($t(17.0)=1.0$, $p = .49$).

\smallskip
\subsection{Usability}
We observe a significant effect of the condition on usability as measured by the SUS, $F(2,34)=23.18$, $p < .001$. Both the PC and the TLA interface were rated as having a high usability (SUS score around 80) significantly outperforming the Cartesian interface (PC-CC: $t(17.0)=4.0$, $p < .001$, TLA-CC: $t(17.0)=-6.0$, $p < .001$). However, we do not observe a significant difference of usability between the PC and the TLA interfaces ($t(17.0)=-1.0$, $p = .69$).

\smallskip
\subsection{Preference}
When asked which methods they preferred, 14 participants replied they preferred the TLA method, three preferred the PC method and one the CC method. Using one-sample binomial test, we measure a significant preference for our TLA method (95\% Adjusted Wald Confidence Interval is (54.24\%, 91.54\%), preference TLA $>$ 33\% with $p < .001$).

\smallskip
\subsection{Observations \& Feedback from Participants}
In addition of our quantitative metrics, we made a number of anecdotal observation during the study and the following semi-structured interview. First, the two main justifications for participants' preference of the TLA interface were the ability to queue actions and the visualization (the two design principles not supported by the PC interface):
\begin{quote}
    ``[TLA was] by far the best, because you could do so many tasks at once, and it was just really intuitive to figure out, okay, this is what it's gonna do''
    
    ``I like that little line thing which would show up on positions, so you could determine like initial position and the final position [...] without having to remember numbers''
    
    ``Being able to angle the jaws, and have visual reference for that, was really useful'' 
\end{quote} 

Some participants used the periods of autonomy of the TLA interface to perform secondary actions, e.g., drinking water or even as one participant did, sending a message to a friend. Combined to  our quantitative results showing that the authoring interface frees longer periods of time to the operators, these observations provide anecdotal evidence that interfaces similar to TLA could help operators perform secondary action. However, as our study did not assess such an hypothesis, future work should confirm it.

Some participants were slightly confused by the different modalities used and monitored the game plan to understand how their inputs were parsed:
\begin{quote}
    ``[TLA] also gave you that menu of like the order that you were going. I feel like that was really helpful''
\end{quote}

Even though many participants qualified the PC interface of being very simple (almost too simple for some), participants still had to follow the progress of their series of action which can be complicated. For example, the screw task requires four repetitions of a loosen, a pick, a reset and a place action. Some participants in the study lost the track of which action was done, and for example forgot to loosen a screw, or did it twice. Some participants felt annoyed to have to re-specify each action every time: 
\begin{quote}
    ``[With TLA] you could I guess perform multiple tasks at once, you didn't have to click on it every single times'' 
    
    ``[PC] it took more time, still because that you had to unscrew it, and then you had to pick it up, and then you had to move it and place it''
\end{quote}
Being able to program the robots to sequence actions when having to repeat them over multiple objects allowed operators to keep track more easily of the progress in the task without having to keep in memory which actions were already executed.

Additionally, due to the lower granularity of control in CC and PC participants reported difficulties to know the distance between the robot's fingers and the table or objects or faced occlusion issues while operating the robot with the PC or CC interfaces. The task-level authoring offered by our system allowed participants to control the robot without facing these two obstacles. 



\section{Discussion}

\subsection{Observations}
Our results provide partial support for our hypotheses. $H_1$ is fully supported (both $P_{1a}$ and $P_{1b}$ are supported). The higher level interfaces performed better than Cartesian Control on all metrics supporting $H_1$.
$H_2$ is partially supported, TLA offered more autonomy than the other methods (supporting $P_{2a}$), and TLA was preferred to the other methods and participants referred to the ability of queuing actions as a reason (supporting $P_{2c}$), however TLA did not reduce the workload compared to PC (failing to support $P_{2b}$). Finally, $H_3$ is also partially supporting. Participants did not achieve a higher score with TLA than with PC and did not rate the usability as higher for TLA (failing to support both $P_{3a}$ and $P_{3b}$), however, participants did prefer TLA over PC and participants referred to the graphical action parameterization as a reason (supporting $P_{3c}$).

Overall, these results show that our design principles partially achieved their goals: the high-level control allowed participants to think at the task level and progress quicker in the tasks. TLA was preferred overall due to the opportunity to create flexible periods of autonomy and the graphical parameterization of actions. This flexible programming horizon allowed participants to specify long periods of autonomy when possible, but also directly select actions when the next step is unclear. Traditionally, robots with more autonomy will require a lower workload at runtime, as the operator does not need to provided inputs when the robot is autonomous. However, such autonomous robots might inflict a higher workload  at design time and require more skills for the operator and more capabilities for the robot. By interleaving exploration, design of short plans, and execution, TLA aims to maintain this low workload both at runtime and design time. Compared to specifying a behavior a priori, allowing the operator to specify commands at runtime allows to solve similar problems, but with simpler robot capabilities (as the operators can perform some sensory analysis) and simpler interface (as the operators does not have to create programs handling every possible situation).

We observed a potential ceiling effect on the usability (a score of 85 on the SUS is defined as excellent usability \citep{brooke2013sus}), and possibly a floor effect on the workload (14 and 11 on the NASA TLX are very low scores). These effects may have two distinct origins. Either our study was not sufficiently challenging for our operators, or our action sequencing principle did allow participants to obtain capabilities closer to programming (through the scheduling of action, automatic generalization of a set of actions to a group of objects etc.), which may have increased the complexity of the interface, but our graphical specification principle balanced this added complexity to maintained a low workload and high usability. Due to time constraints and study complexity, we could not explore individually the impact of each axis, which prevents us to identify the root cause of this effect. Future work should investigate more precisely the situations in which these methods could differ in usability and workload.

Nevertheless, from our study we can confirm that our authoring interface allowed participants to specify longer plans for the robot and streamlined the execution of repeated and composite actions. These two additional benefit might allow operators to perform secondary task, and potentially facilitate extended use (as anecdotally supported by the observation that some participants lost track of their progress in the repeated action and did the same action twice, or forgot whether they unscrewed a bolt already). This additional gain comes at no cost in term of workload and usability, which supports the conclusion that our design principles allowed the interface to be usable with limited training while incorporating additional programming capabilities. Future work should evaluate whether such increase in autonomy could allow operators to perform secondary tasks in practice and how such programming capabilities could be used by operators.

\smallskip

\subsection{Limitations}

Our approach suffers from a number of limitations that we plan to address in future work. 
A key limitation is that the high-level interface requires the specific primitives and actions to be pre-determined and pre-programmed. 
Extending the set of operations to support a broader range of tasks may create challenges in helping the user understand the range of options.
Allowing users to specify actions that are not in the interfaces ``vocabulary'' is challenging, as this requires detailed specification that often must consider low-level control issues such as compliance. This issue is common in authoring---for example, teach pendants are also intrinsically limited in the robot's capabilities they expose and more complex uses often require coding. Additionally, our system relied on robust actions and we did not explore how to recover from failures when executing actions. We identify four ways such action failures could be handled. First, actions could be made more robust by integrating replanning strategies (e.g., planning a new grasp pose after a failed grasp). Second, high-level actions could take more parameters after a failure (e.g., specifying a full 6D grasp pose if the default one did not work). Third, the operator could provide additional runtime inputs to address small trajectory errors \citep{hagenow2021corrective}. And finally, the user could change the control mode for such infrequent event (e.g., temporarily using direct control instead of TLA).

The evaluation of our approach also has a number of limitations. For example, the study considered relatively simple tasks, used a mostly male population, and our population was not total novice, but had some experience with programming. Additionally, due to time constraints, we could not explore every single design axis individually. Future work should involve ablation studies, where the specific impact of design principles are evaluated, explore interactions in real environments, and use operators from the targeted population (family member controlling a robot in a home-assistant scenario or workers in a factory). Finally, future work should also explicitly explore the impact of latency when performing task-level authoring, especially compared to more direct control. We plan to address such limitations in future work.

\smallskip

\subsection{Implications}

Results from our evaluation lead a number of implications. Centrally, the use of task-level authoring seems to be an interesting trade-off, allowing for sufficient programming to gain the advantages of asynchronous control (i.e., programming longer periods of autonomy for the robot and leading to longer and better quality idle time, offloading some tasks following to the robot), yet having the programming be simple enough that it can be used during the interaction with little training. The approach affords an interface design that combines exploration, specification, and monitoring in a single view. The specific interface provides other general lessons. First, our work expands on the ideas of using higher-level controls to enable effective teleoperation interfaces. While prior systems have shown point-and-click interfaces \citep{schmaus2019knowledge}, ours expands the concept to accomplish longer autonomous behavior. Second, by connecting these higher-level controls in a paradigm where exploration and manipulation are interleaved, we can create single-view interfaces that are usable in more complex scenarios. Third, our work extends prior see-through interfaces with camera control, allowing them to work in more environments. Fourth, our work shows the potential of asynchronous interfaces by improving the amount and duration of the offered periods of autonomy. By allowing the user to quickly specify longer plans, they gain opportunities for idle time, potentially freeing them to perform other tasks during execution. Finally, by demonstrating effective telemanipulation only using consumer interfaces shows that remote robot operation is possible for novice users---even at distances of many time zones.

\section{Conclusion}
In this paper, we explored the design of interfaces for remote control of a robotic arm by novice users. 
Our design considers the key goals of teleoperation interfaces: allowing remote novice operators to analyze the robot's environment and specify robot behavior appropriate to the situation.
To address these challenges for scenarios with novice users and standard input devices we adopted a task-level authoring approach. 
The approach allowed for the design of an interface that interleaves exploration and planning, allowing us to utilize both direct control (more intuitive interface and benefiting from the human knowledge more frequently) and asynchronous control (robustness to communications issues and increased idle time for the operator). Our interface uses graphical overlays on a video feed of the environment to provide for simple exploration, specification of operations, and sequencing of commands into short programs. We evaluated a prototype system in an 18-participant study which showed that our interface allowed users with some familiarity with programming to (1) operate the robot remotely to gain awareness about the environment, (2) perform manipulation of the workspace, and (3) use the scheduling of actions to free long periods of idle times that might be used to perform secondary tasks. Furthermore, our interface was largely preferred compared to two other simpler interfaces. 

Our work adds a new tool to the existing library of teleoperation approaches and demonstrates that task-level authoring is a powerful method to allow non-experts to remotely create short periods of autonomy for robots while allowing them to explore the robot's environment. 



\section*{Conflict of Interest Statement}

The authors declare that the research was conducted in the absence of any commercial or financial relationships that could be construed as a potential conflict of interest.



\section*{Funding}
    This work was supported by a NASA University Leadership Initiative (ULI) grant awarded to the UW-Madison and The Boeing Company (Cooperative Agreement \# 80NSSC19M0124).

\bibliographystyle{frontiersinSCNS_ENG_HUMS}
\bibliography{biblio}


\end{document}